\documentclass{article}

\PassOptionsToPackage{numbers, compress}{natbib}

\usepackage[preprint]{neurips_2026}

\newif\ifanonymous
\anonymousfalse 

\usepackage[utf8]{inputenc}
\usepackage[T1]{fontenc}
\usepackage{hyperref}
\usepackage{url}
\usepackage{booktabs}
\usepackage{tabularx}
\usepackage{amsfonts}
\usepackage{amsmath}
\usepackage{amssymb}
\usepackage{nicefrac}
\usepackage{microtype}
\usepackage{xcolor}
\usepackage{colortbl}
\usepackage{graphicx}
\usepackage{multirow}
\usepackage{subcaption}
\usepackage{enumitem}
\usepackage{placeins}

\definecolor{linkblue}{HTML}{1B4F72}
\definecolor{oslerrow}{HTML}{EAF2F8}
\hypersetup{
  colorlinks=true,
  linkcolor=linkblue,
  citecolor=linkblue,
  urlcolor=linkblue
}

\setlist[itemize]{nosep, leftmargin=1.35em, topsep=1.5pt}
\setlist[enumerate]{nosep, leftmargin=1.45em, topsep=1.5pt}
\newcommand{\releaseurl}[2]{%
  \ifanonymous
    \textit{[anonymized for review]}%
  \else
    \href{#2}{\texttt{#1}}%
  \fi
}

\newcommand{\figinclude}[3][width=\linewidth]{%
  \IfFileExists{#2}{%
    \includegraphics[#1]{#2}%
  }{%
    \fbox{\parbox{0.96\linewidth}{\centering\vspace{1.6cm}%
      \small\textit{Figure pending upload: \texttt{#2}}\\[4pt]%
      \footnotesize #3%
    \vspace{1.6cm}}}%
  }%
}

\title{Osler: Open Medical Decision Models and MedDecide-Bench}

\author{%
  Abhinand Balachandran\thanks{Work conducted independently outside of coursework.} \\
  Georgia Institute of Technology \\
  \texttt{abalachandran7@gatech.edu}, \texttt{abhinand5899@gmail.com}
}

\begin{document}

\maketitle

\begin{abstract}
Routine medical workflows---verifying whether a discharge summary supports a documented diagnosis, assigning an MS-DRG, screening trial eligibility, or flagging drug warnings---are bounded decisions over noisy text rather than prose generation. Decision models score candidate options in tens of milliseconds, yet existing open models stumble on real EHR documentation, cap out at ten options, or output poorly calibrated confidence scores. Fine-tuning them on clinical records hits two persistent failures: general medical QA drops by 2--4 points (even with 27B--120B teacher distillation), and accuracy collapses past 26 options when option codes shift from one to two letters. We release \textbf{Osler-9B} (8.39B, non-generative) and \textbf{Osler-4B} (4.55B), Apache-2.0 medical decision models trained with per-item gradient weighting, dynamic option-count augmentation, format-matched KL regularization to the base model, dosed teacher distillation, and dev-only temperature scaling, alongside \textbf{MedDecide-Bench}---a unified 14-model, 9-set evaluation (35,769 items) featuring a CC0-1.0 post-cutoff fresh tier and live leaderboard. Across five headline sets, Osler-9B ranks first overall (78.0\% macro accuracy vs.\ 76.0\% for 27B \texttt{pplx-27B} and 72.6\% for 9B \texttt{elephant-1}), ranks first on fresh post-cutoff decisions (90.1\%) and External Panel calibration (ECE 0.023), holds statistical parity with its base on external QA, and leads \texttt{pplx-27B} by +9.6 points on clinical documentation and EHR tasks at 2.7$\times$ lower latency (33.7\,ms) and 3.1$\times$ less VRAM (17.3\,GB). Osler-4B ranks third overall (75.3\%) in 9.6\,GB VRAM (26.5\,ms), strictly dominating \texttt{MedDecider-4B} across every headline set with a candid $\sim$1-point general-QA trade-off against its base. A local MIMIC-IV privacy audit clears visited-member membership inference ($\Delta\text{AUC}\leq 0.05$), zero verbatim extractions on Osler-4B (0/300 notes), and architectural non-generativity on Osler-9B.
\end{abstract}

\section{Introduction}
\label{sec:intro}

Much of day-to-day clinical practice and medical informatics is not writing prose; it is making structured decisions from messy clinical text. Checking whether a discharge summary supports a documented sepsis diagnosis, assigning an MS-DRG or emergency-department acuity level, screening trial eligibility, or flagging a drug-label warning all share one interface: given a document and a typed question, assign a calibrated probability to every allowed option. Running these checks at hospital scale with autoregressive chain-of-thought LLMs is slow (seconds per chart), brittle to parse, poorly calibrated, and introduces free-text hallucination and patient-text emission risk.

Decision models sidestep those bottlenecks by scoring all offered options in tens of milliseconds. General decision models such as \texttt{pplx-decider}~\citep{perplexity2026pplxdecider}, \texttt{d3-flash}~\citep{vllmsr2026d3flash}, and \texttt{elephant-1}~\citep{baselabs2026elephant1} already perform well on medical exams~\citep{jin2021medqa,pal2022medmcqa}, yet they struggle badly on real clinical notes and electronic health record (EHR) workflows. Early medical decision models such as \texttt{MedDecider-4B/9B}~\citep{lionai2026meddecider} hard-cap prompts at 10 options---dropping differential-diagnosis and specialty-coding tasks with 11--255 labels---and remain uncalibrated for selective triage.

Closing this gap through clinical fine-tuning turned out to be much harder than expected. Across eleven iterative training cycles preceding our final models, two failure modes appeared again and again:
\begin{enumerate}
  \item \textbf{Out-of-domain (OOD) forgetting that distillation alone did not fix.} Naive fine-tuning on real clinical notes and structured MIMIC-IV~\citep{johnson2023mimic} EHR tasks drove large clinical-record gains but pulled accuracy on an external 7-dataset clinical QA panel down by 2--4 points. Surprisingly, adding heavy knowledge distillation~\citep{hinton2015distilling} from 27B--120B teachers restored medical exam scores while leaving external-panel accuracy 1--4 points below the starting model.
  \item \textbf{Two-letter option-code collapse.} Decision readouts supporting up to 255 options assign single letters (\texttt{A}--\texttt{Z}) to options 1--26 and two-letter codes (\texttt{AA}--\texttt{IV}) beyond that. Because routine clinical templates rarely exceed 10 options, standard fine-tuning starved the two-letter readout rows of gradient updates and collapsed accuracy on high-cardinality differential diagnosis (DDXPlus~\citep{tchango2022ddxplus}) to 0.4\%.
\end{enumerate}

\begin{figure}[t]
  \centering
  \figinclude[width=0.88\linewidth]{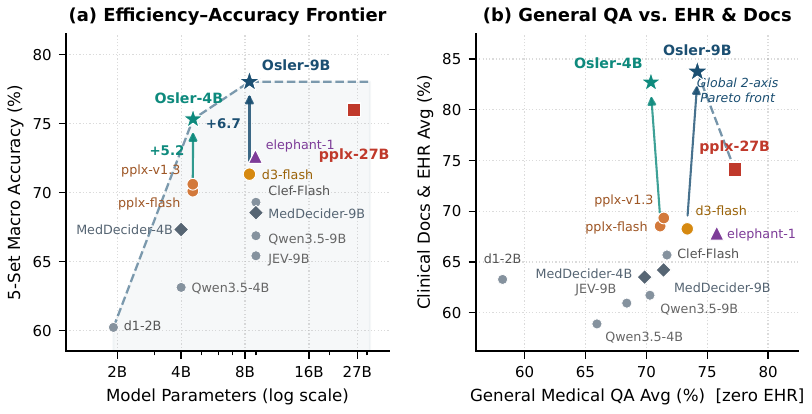}{Hero Pareto frontier: (a) Model size vs.\ 5-set macro accuracy; (b) Global 2-axis capability frontier.}
  \caption{\textbf{Accuracy--efficiency and capability frontiers across MedDecide-Bench (14 models, 33,153 headline items).} \textit{(a)} Both Osler-4B (4.55B, 26.5\,ms, 9.6\,GB) and Osler-9B (8.39B, 33.7\,ms, 17.3\,GB) sit on the efficiency frontier; Osler-9B beats 27B \texttt{pplx-27B} (78.0\% vs.\ 76.0\%) at 2.7$\times$ lower latency and 3.1$\times$ less VRAM. \textit{(b)} On the global two-axis plane (zero-EHR General QA vs.\ Clinical Docs \& EHR), Osler-9B is the sole $\leq$10B model on the Pareto front alongside \texttt{pplx-27B}.}
  \label{fig:pareto}
\end{figure}

We release \textbf{Osler-9B} (8.39B) and \textbf{Osler-4B} (4.55B) together with \textbf{MedDecide-Bench}, an open evaluation suite, interactive clinical vignette explorer, and live leaderboard.\footnote{\raggedright\textbf{Open Artifacts:} Site \& Vignettes: \releaseurl{abhinand5.github.io/MedDecide}{https://abhinand5.github.io/MedDecide}; Weights: \releaseurl{abhinand/Osler-9B}{https://huggingface.co/abhinand/Osler-9B}, \releaseurl{abhinand/Osler-4B}{https://huggingface.co/abhinand/Osler-4B}; Benchmark: \releaseurl{abhinand/MedDecide-Bench}{https://huggingface.co/datasets/abhinand/MedDecide-Bench}; Leaderboard: \releaseurl{abhinand/MedDecide-Leaderboard}{https://huggingface.co/spaces/abhinand/MedDecide-Leaderboard}; Code: \releaseurl{github.com/abhinand5/MedDecide}{https://github.com/abhinand5/MedDecide}.}
Because calibrated medical decision models lack a pre-existing standard benchmark, a fair concern is whether strong EHR scores simply reflect home-field templates. We therefore separate our evidence into two layers. On \emph{strictly external sets containing zero EHR templates}, Osler-9B remains on the size--accuracy frontier for general medical QA, ranks first of all 14 models on fresh post-cutoff decisions (\texttt{confirm\_general}, dated $\geq$2026-03-01), ranks first (with Osler-4B second) on independent MedNLI~\citep{romanov2018mednli}, and achieves the lowest External Panel calibration error on the board (ECE 0.023). Across all five headline sets, Osler-9B is the only $\leq$10B model on the global two-axis front alongside the 27B ceiling (Figure~\ref{fig:pareto}b).

\paragraph{Contributions.}
\begin{itemize}
  \item \textbf{Open Calibrated Medical Decision Models:} Standalone merged Apache-2.0 weights (\textbf{Osler-9B}, \textbf{Osler-4B}) with bundled runtimes and dev-fitted calibration. Osler-9B carries no language-modeling head (architecturally non-generative), and Osler-4B runs in 9.6\,GB VRAM at 26.5\,ms.
  \item \textbf{Anti-Forgetting Recipe \& Empirical Lessons:} A candid account of why pure 27B--120B teacher distillation still degraded external-panel accuracy, paired with a matched training recipe (per-item weighting, option-count augmentation, format-matched KL-to-base, dosed distillation, and dev-only temperature scaling~\citep{guo2017calibration}) that delivers +14--17 point EHR gains with zero OOD tax at 9B (and a documented $\sim$1-point trade-off at 4B).
  \item \textbf{MedDecide-Bench \& Live Leaderboard:} 35,769 frozen items across 9 sets---including a CC0-1.0 post-cutoff structured-gold tier (5,988 items)---evaluated uniformly across 14 models with paired bootstrap CIs, readout-health checks, calibration, selective triage curves, and standardized latency/VRAM benchmarks.
  \item \textbf{Local MIMIC-IV Privacy Audit:} Visited-member membership inference~\citep{shokri2017membership} calibrated against zero-shot temporal controls ($\Delta\text{AUC}\leq 0.05$), 300-note verbatim probing~\citep{carlini2021extracting} (0/300 on 4B), and non-generativity verification on 9B.
\end{itemize}

\section{Typed Clinical Decisions and MedDecide-Bench}
\label{sec:bench}

\paragraph{Typed decision interface.}
Given a document state $x$, a question $q$, and an ordered list of $K$ candidate options $\mathcal{O} = (o_1, \dots, o_K)$ ($2 \le K \le 255$), a decision model outputs a probability distribution $p = f_\theta(x, q, \mathcal{O}) \in \Delta^{K-1}$ in one forward pass without generating text. Every item in MedDecide-Bench uses one executable schema spanning three question types: \texttt{noul} (binary \texttt{yes}/\texttt{no}, returning $p(\text{yes})$), \texttt{choice} ($2 \le K \le 255$ options keyed \texttt{A}--\texttt{Z} then \texttt{AA}--\texttt{IV}), and \texttt{score} ($2 \le K \le 10$ ascending ordinal levels keyed \texttt{1}--\texttt{K}, returning both the level distribution and expected score). Each item receives a deterministic 16-hex SHA-256 identifier; answer-bearing fields are stripped from $x$ by construction, and gold labels come strictly from structured source fields or official benchmark keys---never from an LLM judge.

\begin{table}[t]
  \centering
  \caption{\textbf{Overview of MedDecide-Bench (35,769 frozen items across 9 sets).} Headline rankings use five primary sets (33,153 items): strictly external \emph{General Medical QA} (zero EHR templates) and \emph{Clinical Docs \& EHR}. All test splits are strictly evaluation-only.}
  \label{tab:bench}
  \footnotesize
  \setlength{\tabcolsep}{3.8pt}
  \begin{tabularx}{\linewidth}{@{}llrrX@{}}
    \toprule
    \textbf{Set} & \textbf{Role} & \textbf{Items} & \textbf{Tmpl.} & \textbf{Sources \& Construction} \\
    \midrule
    \multicolumn{5}{@{}l}{\textit{Headline --- General Medical QA (strictly external, zero EHR templates)}} \\
    \texttt{panel} & OOD General & 4,353 & 7 & 7 public QA sets (MedConceptsQA, MedXpertQA, etc.) \\
    \texttt{knowledge} & OOD Exams & 5,456 & 2 & MedMCQA val (4,183) + MedQA USMLE test (1,273) \\
    \texttt{confirm\_general} & Fresh Tier-2 & 5,988 & 12 & Post-cutoff ($\geq$2026-03-01) CT.gov, openFDA, PubMed \\
    \midrule
    \multicolumn{5}{@{}l}{\textit{Headline --- Clinical Documentation \& EHR Tasks}} \\
    \texttt{mr\_verdict} & Clinical Docs & 13,103 & 7 & PMC-Patients, CodiEsp, MTSamples, Synthea, ICD notes \\
    \texttt{mimic\_suite} & MIMIC EHR & 4,253 & 9 & MIMIC-IV ($\geq$2014 split): DRG, ED acuity/disp., ICD, eMAR \\
    \midrule
    \multicolumn{5}{@{}l}{\textit{Additional \& Held-Out Diagnostic Sets}} \\
    \texttt{mednli\_test} & Clinical NLI & 1,422 & 1 & Official MedNLI test split~\citep{romanov2018mednli} \\
    \texttt{pairs\_test} & Minimal Pairs & 500 & 1 & Label-flipping clinically edited MIMIC-IV minimal pairs \\
    \texttt{mr\_heldout\_test} & Held-Out Docs & 584 & 3 & Three clinical-doc templates excluded from all training \\
    \texttt{mimic\_heldout} & Held-Out EHR & 110 & 3 & Held-out heart-failure ICD family \& AKI indicator tasks \\
    \midrule
    \textbf{Total} & \textbf{9 frozen sets} & \textbf{35,769} & \textbf{45} & \textbf{5 headline sets (33,153) + 4 diagnostic sets (2,616)} \\
    \bottomrule
  \end{tabularx}
\end{table}

\paragraph{Nine frozen sets (35,769 items).}
MedDecide-Bench pairs external benchmarks, a contamination-resistant fresh tier, and real clinical-record tasks (Table~\ref{tab:bench}). \emph{General Medical QA} (15,797 items) combines \texttt{panel}---seven unseen public datasets spanning MedConceptsQA~\citep{benshoham2024medconceptsqa}, MedExpQA~\citep{alonso2024medexpqa}, MedExQA~\citep{kim2024medexqa}, Medical Question Pairs~\citep{mccreery2020mqp}, MedXpertQA~\citep{zuo2025medxpertqa}, MMLU-Pro Health~\citep{wang2024mmlupro}, and Symptom-to-Diagnosis (with verified zero exact or 13-gram training overlap on MedConceptsQA)---\texttt{knowledge} (MedQA~\citep{jin2021medqa}, MedMCQA~\citep{pal2022medmcqa}), and our public \textbf{CC0-1.0 fresh tier} \texttt{confirm\_general} (5,988 structured-gold items from ClinicalTrials.gov, openFDA, and PubMed records dated $\geq$2026-03-01, split by record hash). \emph{Clinical Docs \& EHR} (17,356 items) spans \texttt{mr\_verdict} (CodiEsp~\citep{miranda2020codiesp}, PMC-Patients~\citep{zhao2023pmcpatients}, 11--26-option MTSamples, Synthea~\citep{walonoski2018synthea}, and public/generated ICD coding notes) and \texttt{mimic\_suite} (nine structured clinical-documentation templates from late-anchor $\geq$2014 patient-disjoint MIMIC-IV 3.1 / Note 2.2 / ED~\citep{johnson2023mimic}, processed strictly locally). Four diagnostic sets (2,616 items) probe clinical NLI, label-flipping minimal edits, and unseen-template failure modes.

\paragraph{Unified 14-model protocol and safeguards.}
We score 14 open models uniformly on the same frozen item lists: Osler-9B/4B, their zero-shot bases (\texttt{d3-flash}~\citep{vllmsr2026d3flash}, \texttt{pplx-flash}~\citep{perplexity2026pplxdecider}), medical decision models (\texttt{MedDecider-9B/4B}~\citep{lionai2026meddecider}), general decision models (\texttt{elephant-1}~\citep{baselabs2026elephant1}, \texttt{pplx-flash-v1.3}, \texttt{Clef-Flash}~\citep{cloudflare2026clef}, \texttt{JEV-9B}~\citep{autotrust2026jev}, \texttt{d1-2B}~\citep{liquidai2026d1}), raw backbones (\texttt{Qwen3.5-9B/4B}~\citep{qwen2026qwen35}), and a 27B ceiling (\texttt{pplx-27B}). All 14 models have 100\% headline coverage (missing diagnostic cells are never imputed). Every metric is recomputed from saved per-item probabilities with 2,000-resample paired bootstrap 95\% CIs, multi-class Brier~\citep{brier1950verification}, and 15-bin top-label ECE. Two evaluation safeguards prevent silent artifacts: (1) \textbf{readout-health gates} flag zero-shot label-mass failures, constant-answer collapse, and below-chance template accuracy as \texttt{READOUT\_FAIL}; and (2) \textbf{paired $\leq$10-option intersections} ensure fair comparison against \texttt{MedDecider}'s hard 10-option cap (\texttt{JEV-9B} caps at 16). The public tier and live Hugging Face Space accept community PRs alongside PhysioNet-DUA private evaluations.

\section{Osler Architecture and Anti-Forgetting Training Recipe}
\label{sec:method}

\paragraph{Architecture and attention-only adaptation.}
In early head-geometry comparisons at 0.8B and 4B, pointer heads over option spans, joint cross-attention schema heads, and a 255-row option-code readout all landed within 0.6 accuracy points of one another. Because data balance, option-cardinality coverage, and regularization dominated readout geometry, we adopt the 255-row option-code readout (initialized from base code embeddings and softmax-normalized strictly over offered options) across two sizes:
\begin{itemize}
  \item \textbf{Osler-9B} (8.39B; built on noncausal \texttt{vllm-sr/d3-flash}~\citep{vllmsr2026d3flash}): Attention-only LoRA~\citep{hu2022lora} ($r=32, \alpha=32$) plus a $[255, 4096]$ readout at the final prompt position. \textbf{Osler-9B carries no \texttt{lm\_head}} (39.8\,MB trained weights: LoRA and readout only), making free-text generation architecturally impossible.
  \item \textbf{Osler-4B} (4.55B; built on noncausal \texttt{perplexity-ai/pplx-decider-flash-v1.2}~\citep{perplexity2026pplxdecider}): Attention-only LoRA ($r=32, \alpha=32$) plus a $[255, 2560]$ readout at per-question markers. Disabling the adapter restores the untouched base exactly (20/20 token-identical generation).
\end{itemize}
Keeping LoRA restricted to attention projections proved important: in earlier all-layer vs.\ attention-only comparisons, adapting MLP layers degraded medical exam accuracy by 6 points on dev, whereas attention-only LoRA preserved medical knowledge while training 1.8$\times$ faster.

\paragraph{Iterative progression: why simple fine-tuning and pure distillation failed.}
Table~\ref{tab:lineage} records our iterative engineering progression across earlier training runs (note that Table~\ref{tab:lineage} traces historical milestones across changing mixes and backbones rather than a single-factor controlled ablation). Adding real clinical notes and structured MIMIC-IV EHR tasks (Stages~1--2) pushed MIMIC accuracy near 80\%, yet dragged External Panel accuracy down 4.2 points and collapsed DDXPlus two-letter accuracy to 0.4\%. Two mechanisms drove that failure: averaging loss per step let long MIMIC notes absorb $\sim$60\% of gradient updates for 36\% of items, while $\leq$10-option templates starved readout rows 26--254. Adding heavy knowledge distillation~\citep{hinton2015distilling} from 27B or 120B-class teachers (Stages~3--5) restored or boosted medical exam scores, yet still left External Panel accuracy 1.2--3.8 points below each starting checkpoint.

\begin{table}[t]
  \centering
  \caption{\textbf{Iterative empirical progression across training stages (historical runs, not a controlled single-factor ablation).} Adding MIMIC lifted clinical EHR scores but degraded OOD \texttt{panel} ($-$4.2 pts) and collapsed two-letter codes; even 27B--120B teacher distillation left negative panel deltas until format-matched KL-to-base, per-item weighting, and option-count augmentation were combined at 9B.}
  \label{tab:lineage}
  \footnotesize
  \setlength{\tabcolsep}{3.4pt}
  \begin{tabularx}{\linewidth}{@{}llXrrrrr@{}}
    \toprule
    \textbf{Stage} & \textbf{Size} & \textbf{Key Recipe Change} & \textbf{Panel} & \textbf{$\Delta$Base} & \textbf{Know.} & \textbf{Docs} & \textbf{MIMIC} \\
    \midrule
    Stage 1 & 4B & Real notes + \texttt{pplx-flash} base & 0.5343 & $-$0.0198 & 0.6752 & 0.8963 & 0.5551 \\
    Stage 2 & 4B & + MIMIC EHR tasks (2-letter collapse) & 0.5254 & $-$0.0421 & 0.6765 & 0.8862 & \textbf{0.7947} \\
    Stage 3 & 4B & + 385k \texttt{pplx-27B} distillation & 0.5360 & $-$0.0245 & 0.6820 & 0.8950 & 0.7703 \\
    Stage 4 & 4/9B & 120B-class hosted teacher (495k pool) & 0.5596 & $-$0.0115 & 0.6837 & 0.8960 & 0.7682 \\
    Stage 5 & 9B & 120B-class local teacher (200k pool) & 0.6210 & $-$0.0383 & 0.7620 & 0.7084 & --- \\
    \midrule
    \rowcolor{oslerrow}
    \textbf{Final} & \textbf{4B} & Full recipe (KL-to-base + weighting + aug.) & 0.5502 & $-$0.0113 & 0.6728 & 0.8941 & 0.7597 \\
    \rowcolor{oslerrow}
    \textbf{Final} & \textbf{9B} & Full recipe (KL-to-base + weighting + aug.) & \textbf{0.5927} & \textbf{+0.0028} & \textbf{0.7322} & \textbf{0.9016} & 0.7733 \\
    \bottomrule
  \end{tabularx}
\end{table}

\paragraph{Final training recipe and dev-only calibration.}
Both Osler-9B and Osler-4B train for 63,304 token-capped steps on our private 1.53M-row training corpus (61.2\% \texttt{choice}, 28.5\% \texttt{noul}, 10.3\% \texttt{score}; 35.4\% local pre-2014 MIMIC-IV; 28.7\% general teacher-labelled anchors; and 69,967 auxiliary format-matched KL rows labelled locally by \texttt{pplx-27B}; full composition in Appendix~\ref{app:mix}):
\begin{enumerate}
  \item \textbf{Per-item weighting \& token-capped batching:} Each step caps batches at 8 items and 8,192 padded tokens and scales post-clip gradients by $n_{\text{items}}/8$, stopping long single-item MIMIC batches from overpowering short QA batches. Each run visits $\sim$15.5\% of the corpus ($\sim$235k rows; Osler-4B completes in 14.1\,h on one RTX PRO 6000 at 57.1\,GB peak VRAM).
  \item \textbf{Dynamic option-count augmentation:} Eligible clinical choice templates (\texttt{mimic\_msdrg\_v1}, \texttt{mimic\_principal\_dx\_v1}) dynamically expand to 8--64 shuffled options during training so two-letter readout rows receive steady gradients. This lifts DDXPlus two-letter accuracy from 0.4\% to 44.5\% (9B) and 48.9\% (4B) while passing all-orders positional health checks (Appendix Figure~\ref{fig:traj}).
  \item \textbf{Format-matched KL-to-base + dosed distillation:} The loss combines per-item cross-entropy, multi-class Brier, dosed teacher KL ($\lambda=3.24$, active on $\sim$33\% of steps over 438k anchors), and format-matched KL-to-base ($0.1\times\mathrm{KL}(p_{\text{base}}\Vert p_\theta)$ on the 70k auxiliary pool, active on $\sim$3.5\% of steps). Post-hoc WiSE-FT weight interpolation~\citep{wortsman2022wiseft} did not beat the trained endpoint on dev and broke positional health at intermediate 9B blends.
  \item \textbf{Dev-only per-type temperature scaling:} After freezing step-63,304 weights, we fit one scalar temperature $T_{\text{qtype}}>0$ per question type by minimizing NLL strictly on our 14,597-row development split~\citep{guo2017calibration}: $p_k^{\text{cal}}\propto\exp(z_k/T_{\text{qtype}})$. Because temperatures are positive scalars, calibration preserves every top-1 prediction (\textbf{0 argmax flips} across all 35,769 test items; exact fitted values in Appendix Table~\ref{tab:calfits}).
\end{enumerate}

\section{Main Results}
\label{sec:results}

Table~\ref{tab:leaderboard} summarizes accuracy across all 14 models and five headline sets (33,153 items, 14/14 full coverage) alongside standardized batch-1 latency and VRAM; extended per-source, diagnostic, and paired $\leq$10-option tables appear in Appendix~\ref{app:details}. Rather than repeating every table cell, we focus below on the four main empirical takeaways.

\begin{table}[t]
  \centering
  \caption{\textbf{Unified 14-model MedDecide-Bench headline leaderboard (33,153 items; 14/14 full coverage).} Best overall in \textbf{bold}; best $\leq$10B \underline{underlined}. Gen.\ = mean(\texttt{panel}, \texttt{knowledge}, \texttt{confirm\_general}); Clin.\ = mean(\texttt{mr\_verdict}, \texttt{mimic\_suite}). $^*$Partial coverage from 10-opt (\texttt{MedDecider}) or 16-opt (\texttt{JEV-9B}) prompt caps (\texttt{panel} 95.1\%, \texttt{confirm\_general} 97.5\%/99.5\%, \texttt{mr\_verdict} 99.2\%). Latency/VRAM on 500 stratified items (1$\times$ RTX PRO 6000, bf16, batch-1); ``---'' = not measured.}
  \label{tab:leaderboard}
  \scriptsize
  \setlength{\tabcolsep}{3.2pt}
  \resizebox{\linewidth}{!}{%
  \begin{tabular}{@{}lr rrr rr rrr rr@{}}
    \toprule
    & & \multicolumn{3}{c}{\textbf{General Medical QA (External)}} & \multicolumn{2}{c}{\textbf{EHR \& Docs}} & \multicolumn{3}{c}{\textbf{Headline Summary}} & \multicolumn{2}{c}{\textbf{Cost (b=1)}} \\
    \cmidrule(lr){3-5}\cmidrule(lr){6-7}\cmidrule(lr){8-10}\cmidrule(lr){11-12}
    \textbf{Model} & \textbf{Params} & \textbf{Panel} & \textbf{Know.} & \textbf{Fresh} & \textbf{Docs} & \textbf{MIMIC} & \textbf{Gen.} & \textbf{Clin.} & \textbf{5-Set} & \textbf{p50} & \textbf{VRAM} \\
     & (B) & (4,353) & (5,456) & (5,988) & (13,103) & (4,253) & \textbf{Avg} & \textbf{Avg} & \textbf{Macro} & (ms) & (GB) \\
    \midrule
    \rowcolor{oslerrow}
    \textbf{Osler-9B} & 8.39 & 0.5927 & 0.7322 & \textbf{\underline{0.9008}} & \textbf{\underline{0.9016}} & \textbf{\underline{0.7733}} & 0.7419 & \textbf{\underline{0.8375}} & \textbf{\underline{0.7801}} & 33.7 & 17.3 \\
    \texttt{pplx-27B} \textit{(ceiling)} & 26.09 & \textbf{0.6510} & \textbf{0.7823} & 0.8843 & 0.8093 & 0.6729 & \textbf{0.7725} & 0.7411 & 0.7600 & 91.0 & 52.9 \\
    \rowcolor{oslerrow}
    \textbf{Osler-4B} & 4.55 & 0.5502 & 0.6728 & 0.8883 & 0.8941 & 0.7597 & 0.7038 & 0.8269 & 0.7530 & 26.5 & 9.6 \\
    \texttt{elephant-1} & 8.95 & \underline{0.6423} & \underline{0.7345} & 0.8968 & 0.7012 & 0.6558 & \underline{0.7579} & 0.6785 & 0.7261 & 75.4 & 19.5 \\
    \texttt{d3-flash} \textit{(9B base)} & 8.39 & 0.5899 & 0.7335 & 0.8774 & 0.7351 & 0.6301 & 0.7336 & 0.6826 & 0.7132 & 33.8 & 17.3 \\
    \texttt{pplx-flash-v1.3} & 4.54 & 0.5594 & 0.6874 & 0.8963 & 0.7610 & 0.6257 & 0.7143 & 0.6933 & 0.7059 & 37.3 & 16.2 \\
    \texttt{pplx-flash} \textit{(4B base)} & 4.54 & 0.5615 & 0.6835 & 0.8889 & 0.7466 & 0.6240 & 0.7113 & 0.6853 & 0.7009 & 25.5 & 9.6 \\
    \texttt{Clef-Flash} & 9.00 & 0.5711 & 0.6864 & 0.8933 & 0.6831 & 0.6306 & 0.7169 & 0.6568 & 0.6929 & --- & --- \\
    \texttt{MedDecider-9B} & 9.00 & 0.5603$^*$ & 0.6997$^*$ & 0.8825$^*$ & 0.6653$^*$ & 0.6189 & 0.7142 & 0.6421 & 0.6853 & --- & --- \\
    \texttt{MedDecider-4B} & 4.00 & 0.5436$^*$ & 0.6703$^*$ & 0.8820$^*$ & 0.6681$^*$ & 0.6019 & 0.6986 & 0.6350 & 0.6732 & --- & --- \\
    \texttt{Qwen3.5-9B} & 9.00 & 0.5419 & 0.6774 & 0.8894 & 0.6315 & 0.6029 & 0.7029 & 0.6172 & 0.6686 & --- & --- \\
    \texttt{JEV-9B} & 9.00 & 0.5127$^*$ & 0.6448 & 0.8943$^*$ & 0.6303 & 0.5885 & 0.6839 & 0.6094 & 0.6541 & --- & --- \\
    \texttt{Qwen3.5-4B} & 4.00 & 0.4831 & 0.6162 & 0.8789 & 0.6176 & 0.5603 & 0.6594 & 0.5889 & 0.6312 & --- & --- \\
    \texttt{d1-2B} & 1.92 & 0.4259 & 0.4555 & 0.8644 & 0.7069 & 0.5587 & 0.5819 & 0.6328 & 0.6023 & \textbf{\underline{11.6}} & \textbf{\underline{3.8}} \\
    \bottomrule
  \end{tabular}}
\end{table}

\paragraph{1. Both Osler sizes sit on the efficiency frontier; Osler-9B leads overall.}
On parameter count, latency, and VRAM vs.\ 5-set macro accuracy, Osler-4B and Osler-9B define the frontier (Figure~\ref{fig:pareto}a). Osler-9B ranks first of all 14 models overall (78.0\% macro accuracy), beating both the 3$\times$ larger \texttt{pplx-27B} by +2.0 points and the slightly larger \texttt{elephant-1} by +5.4 points. On the two-axis plane separating external medical QA from clinical documentation and EHR tasks (Figure~\ref{fig:pareto}b), Osler-9B is the only $\leq$10B model on the global Pareto front alongside \texttt{pplx-27B}. Across all five headline sets among $\leq$10B models, Osler-9B strictly dominates 10 of 12 competitors with positive margins on every set; only its own base \texttt{d3-flash} (a statistical tie on \texttt{knowledge}) and \texttt{elephant-1} (stronger on \texttt{panel}, yet nearly 16 points behind on EHR \& Docs) remain undominated.

\paragraph{2. External general-QA parity proves gains are not an EHR artifact.}
Looking strictly at the three external sets with zero EHR templates, Osler-9B remains on the size--accuracy frontier for general medical QA between 4.5B \texttt{pplx-flash-v1.3} and 9.0B \texttt{elephant-1}. Crucially, it holds full statistical parity with its own \texttt{d3-flash} base on both \texttt{panel} ($+0.3$ pts, 95\% CI $[-0.7, +1.3]$) and \texttt{knowledge} ($-0.1$ pts $[-0.9, +0.6]$), closing the 2--4 point forgetting gap seen in earlier training stages. On the fresh post-cutoff tier (\texttt{confirm\_general}), Osler-9B places \textbf{first of all 14 models} at 90.1\%, beating \texttt{d3-flash} by +2.3 points and \texttt{pplx-27B} by +1.7 points, aided by $>$12-point gains on fresh openFDA boxed-warning verification.

\paragraph{3. Clinical documentation gains come from messy real records, not trivial templates.}
On clinical documentation and MIMIC-IV EHR tasks, Osler-9B leads \texttt{pplx-27B} by +9.2 points on \texttt{mr\_verdict} and +10.0 points on \texttt{mimic\_suite}, and improves over its own base by +14 to +17 points. Per-template breakdowns confirm these gains concentrate on hard real-world records: on Spanish/English clinical ICD coding (CodiEsp), both Osler sizes more than double \texttt{d3-flash} and lead \texttt{pplx-27B} by over 24 points; similar 14--20 point gains over the 27B ceiling appear on MIMIC emergency-department acuity, DRG complication levels, and MS-DRG assignment.

\paragraph{4. Osler-4B is a compact EHR specialist with a candid OOD trade-off.}
Running in 9.6\,GB VRAM, Osler-4B ranks third overall (75.3\%)---beating 9B \texttt{elephant-1} at half the memory and 2.8$\times$ lower latency, and strictly dominating \texttt{MedDecider-4B}, \texttt{Qwen3.5-4B}, and \texttt{d1-2B} across every headline set. Against its own \texttt{pplx-flash} base, Osler-4B adds +14 to +15 points on clinical documentation and EHR tasks while conceding roughly one point on \texttt{panel} and \texttt{knowledge}---a clear capacity trade-off at 4B. Even when restricted to exact paired $\leq$10-option intersections so \texttt{MedDecider} is not penalized for dropped items (Appendix Table~\ref{tab:additional}), both Osler sizes still outperform \texttt{MedDecider} on \texttt{panel} and \texttt{confirm\_general}. On our additional diagnostic sets, Osler-9B and Osler-4B rank first and second of all 14 models on both independent \texttt{mednli\_test} (89.2\% / 89.1\%) and label-flipping MIMIC minimal pairs (\texttt{pairs\_test}, 96.2\% / 95.8\%); on \texttt{mimic\_heldout}, Osler-9B edges its base while trailing \texttt{pplx-27B}, and both models fail one excluded synthetic template in \texttt{mr\_heldout\_test} (Section~\ref{sec:limitations}).

\section{Calibration, Selective Triage, Cardinality, and Latency}
\label{sec:analysis}

\begin{figure}[t]
  \centering
  \figinclude[width=0.88\linewidth]{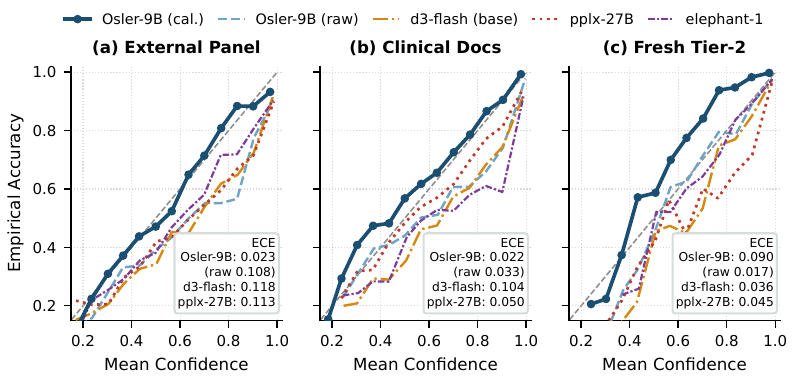}{Three-panel reliability diagrams for External Panel, Clinical Docs, and Fresh Tier-2.}
  \caption{\textbf{Reliability diagrams before and after dev-only temperature scaling.} \textit{(a--b)} Dev temperatures align Osler-9B with the diagonal on OOD External Panel (ECE 0.108 $\to$ \textbf{0.023}, lowest of all 14 models) and Clinical Docs (ECE \textbf{0.022}) with zero argmax flips. \textit{(c)} On higher-accuracy fresh \texttt{confirm\_general}, the same temperatures leave Osler-9B \emph{under-confident} (ECE 0.090 vs.\ raw 0.017).}
  \label{fig:calibration}
\end{figure}

\paragraph{Calibration transfers well to OOD QA, but under-predicts on easier fresh sets.}
Dev-only temperature scaling~\citep{guo2017calibration} cuts calibration error sharply on unseen clinical tasks without touching a single test label (Figure~\ref{fig:calibration}). On External Panel, Osler-9B and Osler-4B rank \textbf{first and second of all 14 models} in calibration (ECE 0.023 and 0.037)---four to five times lower than \texttt{elephant-1} and \texttt{pplx-27B}---and Osler-9B holds similarly tight calibration on clinical documentation (ECE 0.022) and medical exams. Calibration transfer is not universal, however: on fresh \texttt{confirm\_general} (Figure~\ref{fig:calibration}c), where accuracy rises near 90\%, the same dev temperatures over-dampen logits and leave both models visibly under-confident (mean confidence $\sim$80--81\% vs.\ $\sim$89--90\% accuracy).

\begin{figure}[t]
  \centering
  \figinclude[width=0.90\linewidth]{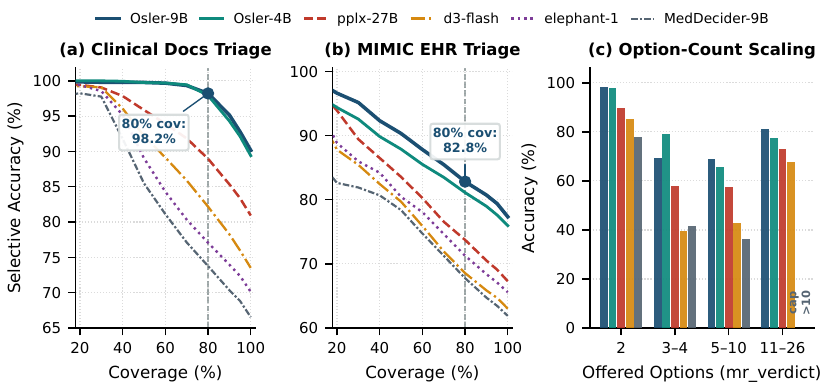}{Selective risk-coverage on Clinical Docs and MIMIC EHR plus option-count cardinality scaling.}
  \caption{\textbf{Selective clinical triage and option-count scaling.} \textit{(a--b)} Routing the lowest-confidence 20\% of items to human review (80\% coverage) pushes Osler-9B to \textbf{98.2\%} accuracy on Clinical Docs (1.8\% error vs.\ 11.0\% for \texttt{pplx-27B}) and \textbf{82.8\%} on MIMIC EHR (+9.1 pts over \texttt{pplx-27B}). \textit{(c)} Osler leads every option bucket on \texttt{mr\_verdict} and scales into 11--26 options (\texttt{mtsamples}, 81.1\% for 9B), where 10-option verbalizers (\texttt{MedDecider}) drop out.}
  \label{fig:selective}
\end{figure}

\paragraph{Selective triage and native high-cardinality scaling.}
Because Osler's confidence scores are well-ordered, abstaining on uncertain cases~\citep{geifman2017selective} unlocks practical human-in-the-loop workflows (Figure~\ref{fig:selective}a--b). Routing just the bottom 20\% of clinical-document predictions to manual review lifts Osler-9B to \textbf{98.2\%} accuracy (a 6$\times$ error reduction vs.\ \texttt{pplx-27B}) and pushes MIMIC EHR accuracy to \textbf{82.8\%}; escalating half the queue reaches 99.8\% on clinical docs and 90.3\% on MIMIC. Across label cardinalities (Figure~\ref{fig:selective}c), Osler avoids \texttt{MedDecider}'s 10-option cliff, leading every bucket on \texttt{mimic\_suite} and scaling cleanly into 11--26-option clinical specialty routing (\texttt{mtsamples}, 81.1\% for 9B vs.\ 73.0\% for \texttt{pplx-27B}) and 22-class symptom diagnosis.

\begin{figure}[t]
  \centering
  \figinclude[width=0.84\linewidth]{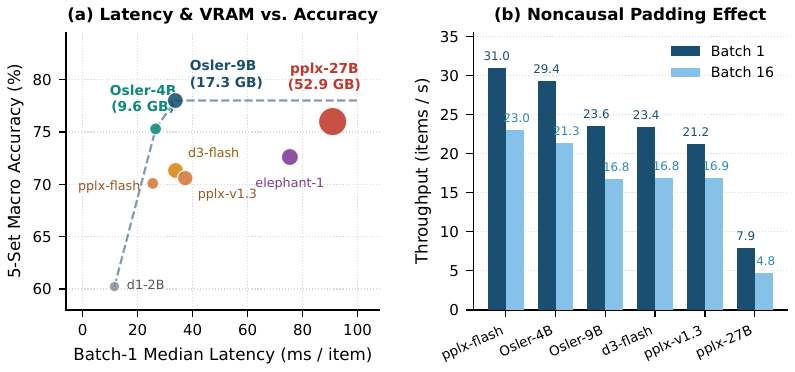}{Batch-1 latency and VRAM vs.\ 5-set macro accuracy and batch-1 vs.\ batch-16 throughput.}
  \caption{\textbf{Latency, VRAM, and noncausal batching (500-item stratified sample, 1$\times$ RTX PRO 6000, bf16).} \textit{(a)} Osler-4B (26.5\,ms, 9.6\,GB) and Osler-9B (33.7\,ms, 17.3\,GB) define the latency--accuracy front, 2.2--2.7$\times$ faster than \texttt{elephant-1} (75.4\,ms) and \texttt{pplx-27B} (91.0\,ms, 52.9\,GB). \textit{(b)} Naive batch-16 throughput is 20--40\% \emph{lower} than batch-1 across all six noncausal models due to long-note padding.}
  \label{fig:efficiency}
\end{figure}

\paragraph{Latency, VRAM, and an architectural serving lesson.}
Benchmarked on the same 500-item stratified sample (half External Panel, half Clinical Docs; 1$\times$ RTX PRO 6000, bf16; Figure~\ref{fig:efficiency}), Osler-4B scores an item in \textbf{26.5\,ms} median using 9.6\,GB VRAM (29.4 items/s), and Osler-9B runs in \textbf{33.7\,ms} using 17.3\,GB (23.6 items/s)---adding virtually zero overhead over its base while running \textbf{2.2$\times$ faster} than \texttt{elephant-1} and \textbf{2.7$\times$ faster with 3.1$\times$ less VRAM} than \texttt{pplx-27B}. Figure~\ref{fig:efficiency}b highlights an unexpected serving pitfall: naive batch-16 throughput is 20--40\% \emph{slower} than batch-1 across every measured noncausal model, because bidirectional full attention pads every item in a batch to the longest clinical note. Production deployments should therefore favor batch-1 continuous scheduling or tight length-bucketed micro-batches.

\section{Local MIMIC-IV Privacy and Memorisation Audit}
\label{sec:privacy}

Roughly a third (35.4\%) of our training corpus derives from MIMIC-IV~\citep{johnson2023mimic}. All MIMIC-IV 3.1, Note 2.2, and ED processing ran strictly on operator-controlled local hardware (zero hosted API calls) under patient-disjoint temporal splits (\texttt{anchor\_year\_group} $\geq 2014$ held out for evaluation), and no MIMIC text, patient IDs, or training rows enter the released weights or repository.

\paragraph{Visited-member audit design and verdict.}
Because our 63,304-step budget visits roughly 15.5\% of the full corpus ($\sim$235k rows per model, with only 8.4\% Jaccard overlap between sizes due to tokenizer-dependent length packing), auditing random training rows would heavily dilute any memorisation signal. We therefore evaluate a targeted \emph{visited-member} set of 2,000 MIMIC items actually seen by \emph{both} models (stratified across all nine \texttt{mimic\_suite} templates) against 2,000 late-anchor non-members verified to have zero content-hash overlap with training (Table~\ref{tab:privacy}):
\begin{enumerate}
  \item \textbf{Membership inference~\citep{shokri2017membership}:} Scoring items by $-\log p(\text{gold})$, even untouched zero-shot bases achieve AUC $\sim$0.54 because pre-2014 train admissions differ slightly in phrasing and case mix from $\geq$2014 test admissions. Subtracting each model's base control isolates training-induced signal: $\Delta\text{AUC}$ is $+0.047$ for Osler-9B and $+0.025$ for Osler-4B, both clearing our pre-registered $\leq 0.05$ release gate. On an unfiltered 4,000-item historical compatibility set, both models sit at chance ($\text{AUC}\approx 0.50$).
  \item \textbf{Verbatim extraction~\citep{carlini2021extracting} \& non-generativity:} Osler-9B carries no language-modeling head (\texttt{lm\_head}), making free-text extraction structurally impossible. Probing Osler-4B's causal export with 128-token prefixes from 300 MIMIC discharge notes yields \textbf{0/300} verbatim matches at $\geq$50, $\geq$25, and $\geq$10 consecutive aligned tokens (maximum run 1 token, median 0).
\end{enumerate}
We stress that this audit is a bounded empirical release check over visited rows rather than a formal differential-privacy guarantee.

\begin{table}[t]
  \centering
  \caption{\textbf{Local MIMIC-IV privacy and memorisation audit.} Visited-member MIA AUC uses $-\log p(\text{gold})$ over 2,000 shared visited members vs.\ 2,000 late-anchor non-members (1,000-resample 95\% CI). Subtracting untouched zero-shot base controls removes temporal admission shift ($\sim$0.54). Both models clear our pre-registered $\Delta\text{AUC}\leq 0.05$ and 0/300 verbatim rules.}
  \label{tab:privacy}
  \footnotesize
  \setlength{\tabcolsep}{3.5pt}
  \resizebox{\linewidth}{!}{%
  \begin{tabular}{@{}lcccccl@{}}
    \toprule
    \textbf{Model} & \textbf{MIA AUC [95\% CI]} & \textbf{Base Control} & \textbf{$\Delta$AUC [95\% CI]} & \textbf{Hist.\ Set} & \textbf{Verbatim ($\geq$50t)} & \textbf{Gate} \\
    \midrule
    \rowcolor{oslerrow}
    \textbf{Osler-9B} & 0.5876 [0.570, 0.605] & 0.5408 (\texttt{d3}) & +0.0468 [+0.032, +0.063] & 0.5018 & N/A (no \texttt{lm\_head}) & \textbf{CLEAR} \\
    \rowcolor{oslerrow}
    \textbf{Osler-4B} & 0.5644 [0.547, 0.581] & 0.5396 (\texttt{pplx}) & +0.0248 [+0.010, +0.040] & 0.5033 & \textbf{0 / 300} (max run 1) & \textbf{CLEAR} \\
    \bottomrule
  \end{tabular}}
\end{table}

\section{Related Work}
\label{sec:related}

\paragraph{Decision models and readout families.}
Open decision models fall into four readout families: single-token letter verbalizers (\texttt{Qwen3.5}~\citep{qwen2026qwen35}, \texttt{MedDecider-4B/9B}~\citep{lionai2026meddecider} capped at 10 options, \texttt{JEV-9B}~\citep{autotrust2026jev} at 16), pointer heads over option spans, joint cross-attention schema heads (\texttt{Clef-Flash}~\citep{cloudflare2026clef}), and 255-row option-code readouts (\texttt{pplx-decider}~\citep{perplexity2026pplxdecider}, \texttt{d3-flash}~\citep{vllmsr2026d3flash}, Osler) alongside slot-based noncausal scoring (\texttt{elephant-1}~\citep{baselabs2026elephant1}) and compact routers (\texttt{d1-2B}~\citep{liquidai2026d1}). Our controlled 0.8B and 4B head ablations found these readout geometries within 0.6 points of one another under matched data; the decisive factors are option-cardinality coverage, loss weighting, and anti-forgetting regularization.

\paragraph{Clinical NLP, medical benchmarks, and adaptation.}
Early clinical encoders such as ClinicalBERT~\citep{alsentzer2019clinicalbert} and recent generative medical LLMs including Med-PaLM~\citep{singhal2023clinical} and MedGemma~\citep{sellergren2025medgemma} have largely been benchmarked on static exam datasets~\citep{jin2021medqa,pal2022medmcqa,romanov2018mednli,wang2024mmlupro,zuo2025medxpertqa}. Real documentation corpora---CodiEsp~\citep{miranda2020codiesp}, PMC-Patients~\citep{zhao2023pmcpatients}, DDXPlus~\citep{tchango2022ddxplus}, Synthea~\citep{walonoski2018synthea}, and MIMIC-IV~\citep{johnson2023mimic}---expose long-context ambiguity and high-cardinality label spaces rarely tested by 4-option exams; MedDecide-Bench bridges both under one executable schema with a post-cutoff fresh tier. While temperature scaling~\citep{guo2017calibration}, Brier scoring~\citep{brier1950verification}, selective classification~\citep{geifman2017selective}, LoRA~\citep{hu2022lora}, distillation~\citep{hinton2015distilling}, and WiSE-FT~\citep{wortsman2022wiseft} are well-studied individually, our results show that neither WiSE-FT nor pure teacher distillation preserves external QA in clinical decision models without attention-only LoRA, format-matched KL-to-base, per-item weighting, and option-count augmentation.

\section{Limitations and Safety Considerations}
\label{sec:limitations}

\begin{enumerate}
  \item \textbf{Multi-step reasoning ceiling and non-generative scope:} Osler-9B ties \texttt{d3-flash} on \texttt{panel}/\texttt{knowledge} and trails 27B \texttt{pplx-27B} on hard multi-step QA (\texttt{medxpertqa\_text}, \texttt{mmlu\_pro\_health}, \texttt{medqa}). Single-pass decision models operate in a 26--34\,ms non-generative regime orthogonal to multi-second chain-of-thought reasoning; they fit grounded verification and calibrated triage rather than deep multi-step diagnostic deduction.
  \item \textbf{4B capacity trade-off:} Osler-4B concedes $\sim$1 point to \texttt{pplx-flash} on \texttt{panel} and \texttt{knowledge} in exchange for +14 to +15 point clinical documentation and EHR gains and strong Brier improvements.
  \item \textbf{Held-out template inversion (\texttt{READOUT\_FAIL}):} On the excluded synthetic template \texttt{synthea\_indicator\_evidence\_v1} in \texttt{mr\_heldout\_test} ($n=514$, 94.4\% majority class), both Osler models drop below chance (35.6\% 9B, 24.5\% 4B) while their bases pass (45.9\% / 59.7\%), underscoring that unseen-template validation remains essential before deployment.
  \item \textbf{Fresh-set calibration shift:} Dev temperatures calibrate \texttt{panel} and \texttt{mr\_verdict} tightly (ECE 0.02--0.04) but leave both models under-confident on higher-accuracy fresh \texttt{confirm\_general} (ECE $\sim$0.09); refitting temperatures on a small target-domain dev split is advised under sharp accuracy shifts.
  \item \textbf{Batching and unmeasured cells:} Naive batch-16 slows noncausal models via long-note padding; standardized latency for \texttt{MedDecider}, \texttt{Clef-Flash}, \texttt{JEV-9B}, and raw \texttt{Qwen3.5} plus 3 of 126 diagnostic cells remain explicitly not measured.
  \item \textbf{Privacy, data access, and clinical scope:} To honor PhysioNet data-use agreements, our training corpus remains private (aggregates, zero-text reconstruction manifests, weights, code, and the CC0-1.0 public benchmark are open). The local MIMIC audit covers visited rows ($\sim$15.5\%) as an empirical release check, not a formal differential-privacy proof. Finally, MIMIC-IV reflects a single academic medical center, and neither model is cleared for unsupervised clinical decision-making.
\end{enumerate}

\section{Conclusion}
\label{sec:conclusion}

We released \textbf{Osler-9B} and \textbf{Osler-4B} (Apache-2.0) together with \textbf{MedDecide-Bench} (35,769 items, 14 models, a CC0-1.0 post-cutoff tier, worked clinical vignettes, and a live leaderboard). By resolving long-note gradient dominance, two-letter option-code collapse, and out-of-domain panel forgetting, our 8.39B non-generative model ranks first overall (78.0\%), leads fresh post-cutoff decisions (90.1\%) and External Panel calibration (ECE 0.023), and beats a 27B ceiling by +9.6 points across clinical documentation and EHR tasks at 2.7$\times$ lower latency and 3.1$\times$ less VRAM---while Osler-4B ranks third overall in 9.6\,GB VRAM. Promising next directions include routing low-confidence single-pass cases to selective chain-of-thought reasoning and prospective multi-center clinical validation.

\bibliographystyle{plainnat}
\bibliography{references}

\newpage
\appendix
\section{Training Corpus Composition, Dynamics, and Dev Calibration}
\label{app:mix}

\begin{figure}[!htb]
  \centering
  \figinclude[width=0.88\linewidth]{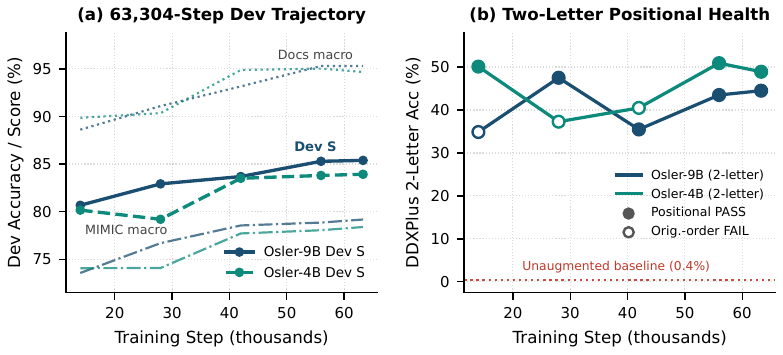}{Two-panel training trajectory: (a) 63,304-step dev S, Docs macro, and MIMIC macro; (b) Two-letter DDXPlus positional health vs.\ unaugmented collapsed baseline (0.4\%).}
  \caption{\textbf{Training dynamics and two-letter positional health over 63,304 steps.} \textit{(a)} Composite dev score $S$, Clinical Docs macro, and MIMIC macro climb steadily across checkpoints for both sizes. \textit{(b)} Dynamic option-count augmentation keeps DDXPlus two-letter accuracy at 44.5\% (9B) and 48.9\% (4B) at step 63,304 (filled = positional PASS; hollow = transient mid-run original-order FAIL on 4B), vs.\ 0.4\% collapse without augmentation.}
  \label{fig:traj}
\end{figure}

Table~\ref{tab:mix} summarizes the aggregate composition of our private 1,526,405-row training corpus (zero raw text, zero patient IDs). Across 63,304 token-capped steps, each model visits $\sim$15.5\% of the corpus (234,251 rows 9B; 235,968 4B). Figure~\ref{fig:traj} shows the 63,304-step dev trajectory alongside DDXPlus two-letter positional health, recovering from the 0.4\% unaugmented collapse to 44.5\% (9B) and 48.9\% (4B). Table~\ref{tab:calfits} details dev-only per-type temperature fits on our 14,597-row development split.

\begin{table}[!htb]
  \centering
  \caption{\textbf{Aggregate composition of the training corpus (1,526,405 rows).} MIMIC-derived rows (539,807; 35.4\%) come strictly from pre-2014 patient-disjoint splits processed locally; automated overlap audit confirms 0/1,000 exact or 13-gram matches against external \texttt{medconceptsqa\_all}.}
  \label{tab:mix}
  \footnotesize
  \setlength{\tabcolsep}{4pt}
  \resizebox{\linewidth}{!}{%
  \begin{tabular}{@{}llrr@{\qquad}lrr@{}}
    \toprule
    \textbf{Dimension} & \textbf{Category} & \textbf{Rows} & \textbf{Share} & \textbf{Top Source} & \textbf{Rows} & \textbf{Share} \\
    \midrule
    \multirow{2}{*}{Pool} & Primary training mix & 1,456,438 & 95.4\% & \texttt{mimic\_iv} (+Note) & 539,807 & 35.4\% \\
     & Auxiliary KL pool & 69,967 & 4.6\% & \texttt{clinicaltrials} & 244,523 & 16.0\% \\
    \midrule
    \multirow{3}{*}{\texttt{qtype}} & \texttt{choice} & 934,743 & 61.2\% & \texttt{pubmed} & 129,132 & 8.5\% \\
     & \texttt{noul} & 434,283 & 28.5\% & \texttt{aqua\_rat} / \texttt{snli} & 131,660 & 8.6\% \\
     & \texttt{score} & 157,379 & 10.3\% & \texttt{openfda} & 61,500 & 4.0\% \\
    \midrule
    Anchors & Teacher-labelled & 438,188 & 28.7\% & \texttt{mr\_pmc} / \texttt{medmcqa} & 103,538 & 6.8\% \\
    \bottomrule
  \end{tabular}}
\end{table}

\begin{table}[!htb]
  \centering
  \caption{\textbf{Dev-only per-\texttt{qtype} temperature calibration fitted on the 14,597-row dev split.} Fitting minimizes NLL strictly on dev with zero test-set tuning and zero argmax flips across all 35,769 test items.}
  \label{tab:calfits}
  \footnotesize
  \setlength{\tabcolsep}{4pt}
  \resizebox{\linewidth}{!}{%
  \begin{tabular}{@{}llrrrrr@{}}
    \toprule
    \textbf{Model} & \textbf{\texttt{qtype}} & \textbf{$n_{\text{dev}}$} & \textbf{$T$} & \textbf{NLL (raw $\to$ cal)} & \textbf{ECE (raw $\to$ cal)} & \textbf{Argmax Flips} \\
    \midrule
    Osler-9B & \texttt{choice} & 9,284 & 1.5274 & 0.6231 $\to$ 0.5672 & 0.0536 $\to$ 0.0265 & 0 \\
    Osler-9B & \texttt{noul}   & 3,680 & 2.3185 & 0.4216 $\to$ 0.3030 & 0.0808 $\to$ 0.0259 & 0 \\
    Osler-9B & \texttt{score}  & 1,633 & 1.7035 & 0.9680 $\to$ 0.8675 & 0.1384 $\to$ 0.0634 & 0 \\
    \rowcolor{oslerrow}
    \textbf{Osler-9B} & \textbf{ALL}    & 14,597 & ---   & 0.6109 $\to$ 0.5342 & 0.0697 $\to$ 0.0192 & 0 \\
    \midrule
    Osler-4B & \texttt{choice} & 9,284 & 1.5222 & 0.6455 $\to$ 0.5884 & --- & 0 \\
    Osler-4B & \texttt{noul}   & 3,680 & 2.3744 & 0.4443 $\to$ 0.3104 & --- & 0 \\
    Osler-4B & \texttt{score}  & 1,633 & 1.5388 & 0.9364 $\to$ 0.8692 & --- & 0 \\
    \bottomrule
  \end{tabular}}
\end{table}

\clearpage
\section{Extended Evaluation Breakdowns}
\label{app:details}

Table~\ref{tab:panel_src} breaks down External Panel accuracy across all seven public OOD datasets for our models and leading baselines; Table~\ref{tab:additional} reports the four additional and held-out sets alongside exact paired $\leq$10-option intersections with \texttt{MedDecider}.

\begin{table}[!htb]
  \centering
  \caption{\textbf{External Panel (\texttt{panel}, 4,353 items) accuracy by source dataset.} ``---'' indicates items dropped by \texttt{MedDecider}'s hard 10-option prompt cap (\texttt{symptom\_to\_diagnosis}, 22 classes).}
  \label{tab:panel_src}
  \scriptsize
  \setlength{\tabcolsep}{3.5pt}
  \resizebox{\linewidth}{!}{%
  \begin{tabular}{@{}lrrrrrrr@{}}
    \toprule
    \textbf{Source Dataset} & \textbf{$n$} & \textbf{Osler-9B} & \textbf{Osler-4B} & \textbf{\texttt{d3-flash}} & \textbf{\texttt{elephant-1}} & \textbf{\texttt{MedDec.-9B}} & \textbf{\texttt{pplx-27B}} \\
    \midrule
    \texttt{medconceptsqa\_all} & 1,000 & 0.5000 & 0.4470 & 0.4590 & 0.4640 & 0.4160 & \textbf{0.5300} \\
    \texttt{medexpqa\_en} & 125 & 0.8080 & 0.7440 & 0.8080 & 0.7760 & 0.7360 & \textbf{0.8960} \\
    \texttt{medexqa\_test} & 940 & 0.8181 & 0.7745 & 0.8277 & 0.8149 & 0.7989 & \textbf{0.8394} \\
    \texttt{medical\_question\_pairs} & 295 & 0.8847 & 0.8610 & 0.8475 & 0.8814 & 0.8746 & \textbf{0.9322} \\
    \texttt{medxpertqa\_text} & 1,000 & 0.2580 & 0.2440 & 0.2820 & \textbf{0.4870} & 0.2600 & 0.3750 \\
    \texttt{mmlu\_pro\_health} & 781 & 0.6978 & 0.6312 & 0.7081 & 0.7362 & 0.6953 & \textbf{0.7759} \\
    \texttt{symptom\_to\_diagnosis} & 212 & 0.6887 & 0.6415 & 0.6840 & \textbf{0.6934} & --- & \textbf{0.6934} \\
    \midrule
    \rowcolor{oslerrow}
    \textbf{Micro Acc / Cal.\ ECE} & 4,353 & 0.5927 / \textbf{0.023} & 0.5502 / 0.037 & 0.5899 / 0.118 & 0.6423 / 0.087 & 0.5603 / 0.046 & \textbf{0.6510} / 0.113 \\
    \bottomrule
  \end{tabular}}
\end{table}

\begin{table}[!htb]
  \centering
  \caption{\textbf{Additional/held-out sets and exact paired $\leq$10-option intersections.} $^\dagger$Indicates \texttt{READOUT\_FAIL} on \texttt{synthea\_indicator\_evidence\_v1} ($n=514$). ``NM'' = not measured.}
  \label{tab:additional}
  \scriptsize
  \setlength{\tabcolsep}{3.8pt}
  \resizebox{\linewidth}{!}{%
  \begin{tabular}{@{}lrrrrrr@{}}
    \toprule
    \textbf{Model} & \textbf{\texttt{mednli\_test}} & \textbf{\texttt{pairs\_test}} & \textbf{\texttt{mimic\_heldout}} & \textbf{\texttt{mr\_heldout}} & \textbf{\texttt{panel} ($\leq$10)} & \textbf{\texttt{confirm} ($\leq$10)} \\
     & (1,422) & (500) & (110) & (584) & (4,141) & (5,838) \\
    \midrule
    \rowcolor{oslerrow}
    \textbf{Osler-9B} & \textbf{0.8917} & \textbf{0.9620} & 0.7000 & 0.4298$^\dagger$ & 0.5878 & \textbf{0.9072} \\
    \rowcolor{oslerrow}
    \textbf{Osler-4B} & 0.8910 & 0.9580 & 0.6727 & 0.3339$^\dagger$ & 0.5455 & 0.8936 \\
    \texttt{pplx-27B} & 0.8748 & 0.9400 & \textbf{0.7818} & \textbf{0.8990} & \textbf{0.6489} & 0.8890 \\
    \texttt{elephant-1} & 0.8713 & 0.9340 & 0.7273 & 0.8219 & 0.6397 & 0.9029 \\
    \texttt{d3-flash} & 0.8629 & 0.9400 & 0.6818 & 0.4914 & 0.5851 & 0.8839 \\
    \texttt{pplx-flash} & 0.8720 & 0.8740 & NM & 0.6130 & 0.5554 & 0.8955 \\
    \texttt{MedDecider-9B} & 0.8425 & 0.7880 & 0.6455 & 0.4024$^\dagger$ & 0.5603 & 0.8825 \\
    \texttt{MedDecider-4B} & 0.8488 & 0.6920 & 0.6636 & 0.5205 & 0.5436 & 0.8820 \\
    \bottomrule
  \end{tabular}}
\end{table}

\FloatBarrier
\section{Worked Clinical Vignettes and Standalone Verification}
\label{app:repro}

\paragraph{Qualitative clinical examples.}
Table~\ref{tab:vignettes} illustrates four freshly measured synthetic clinical vignettes run directly on our standalone merged checkpoints with built-in dev calibration. On ordinal \texttt{score} tasks (e.g., grading warfarin--TMP-SMX bleeding urgency), unanchored single-word scales cluster toward the middle level, so production rubrics should attach explicit clinical thresholds per level. All eight vignettes and an interactive explorer live at \releaseurl{abhinand5.github.io/MedDecide}{https://abhinand5.github.io/MedDecide} and \releaseurl{github.com/abhinand5/MedDecide}{https://github.com/abhinand5/MedDecide}.

\begin{table}[!htb]
  \centering
  \caption{\textbf{Measured clinical vignettes on standalone merged Osler-9B and Osler-4B} (built-in dev calibration, zero generation).}
  \label{tab:vignettes}
  \scriptsize
  \setlength{\tabcolsep}{3.0pt}
  \renewcommand{\arraystretch}{0.95}
  \begin{tabularx}{\linewidth}{@{}lXlll@{}}
    \toprule
    \textbf{Scenario} & \textbf{Clinical State Summary \& Question} & \textbf{Type} & \textbf{Osler-9B} & \textbf{Osler-4B} \\
    \midrule
    Negation / PE & Chest pain after flight; CTA shows \emph{no} PE; normal troponin; resolved on PPI. \textit{Confirmed PE?} & \texttt{noul} & $p(\text{yes})=0.043$ & $p(\text{yes})=0.027$ \\
    EHR Distractor & Inactive childhood asthma, EF 25\%, orthopnea, BNP 2,650, Kerley B, wheeze. \textit{Primary driver?} & \texttt{choice} & \texttt{hfref} (0.997) & \texttt{hfref} (0.992) \\
    Abstention & 1.8\,cm spiculated lung nodule on CT; PET/biopsy pending, no pathology yet. \textit{Confirmed TNM stage?} & \texttt{choice} & \texttt{insuff.} (0.968) & \texttt{insuff.} (0.990) \\
    Minimal Pair & \textit{E.\ coli} bacteremia: (Pos) MAP 56 on pressors, lactate 4.6 vs.\ (Neg) MAP 78, lactate 1.2. \textit{Septic shock?} & \texttt{noul} & \textbf{0.973} / \textbf{0.042} & \textbf{0.973} / \textbf{0.037} \\
    \bottomrule
  \end{tabularx}
\end{table}

\paragraph{Standalone release verification.}
Both checkpoints ship as standalone merged BF16 Apache-2.0 packages (\texttt{AutoModel.from\_pretrained(..., trust\_remote\_code=True)}): \textbf{Osler-9B} (\releaseurl{abhinand/Osler-9B}{https://huggingface.co/abhinand/Osler-9B}, 4 shards + FP32 $[255,4096]$ readout, \texttt{has\_lm\_head = false}) and \textbf{Osler-4B} (\releaseurl{abhinand/Osler-4B}{https://huggingface.co/abhinand/Osler-4B}, 2 shards + $[255,2560]$ readout). Offline checks verified zero-drift adapter reloading ($\max|\Delta p|=0.0$, 20/20 argmax), 5/5 merged BF16 parity ($\max|\Delta p|\leq 0.0111$), and 20/20 token-identical base generation on Osler-4B with adapter disabled. Credentialed PhysioNet users can reconstruct private MIMIC-IV sets deterministically from our published zero-text SHA-256 manifests.

\end{document}